# Modeling trajectories of mental health: challenges and opportunities

**Lauren Erdman[1], Ekansh Sharma[1], Eva Unternährer[2], Shantala Hari Dass[2], Kieran O'Donnell[2], Sara Mostafavi[3], Rachel Edgar[3], Michael Kobor[3], Hélène Gaudreau[4], Michael Meaney[2], Anna Goldenberg[1]**

[1]University of Toronto Department of Science,
[2]McGill University Ludmer Centre for Neuroinformatics and Mental Health,
[3]University of British Columbia Centre for Molecular Medicine and Therapeutics
[4]Douglas Mental Health University Institute

## Abstract

More than two thirds of mental health problems have their onset during childhood or adolescence. Identifying children at risk for mental illness later in life and predicting the type of illness is not easy. We set out to develop a platform to define subtypes of childhood social-emotional development using longitudinal, multifactorial trait-based measures. Subtypes discovered through this study could ultimately advance psychiatric knowledge of the early behavioral signs of mental illness. To this extent we have examined two types of models: latent class mixture models and GP-based models. Our findings indicate that while GP models come close in accuracy of predicting future trajectories, LCMMs predict the trajectories as well in a fraction of the time. Unfortunately, neither of the models are currently accurate enough to lead to immediate clinical impact. The available data related to the development of childhood mental health is often sparse with only a few time points measured and require novel methods with improved efficiency and accuracy.

## 1   Introduction

Mental disorders constitute the largest contributor to the global burden of disease as measured using the disability-adjusted life years index [1]. The most common mental disorders, including attention deficit hyperactive disorder and major depression show a peak age of onset in childhood and adolescence thus derailing the quality of life and productivity of individuals over entire lifetimes. By identifying at risk children at an early age we have an opportunity to intervene and reduce the negative consequences of or prevent many such mental disorders. The challenge is in effectively identifying truly vulnerable children to be able to intervene in a timely manner. Current programs for identifying at risk children are constructed on evidence linking early life adversity, such as poverty or birth outcomes, and the risk for mental illness. These factors predict mental illness at the level of the population, but are ineffective at the level of the individual due to the considerable variability in outcomes: many children born early, small, or into poverty are healthy and productive.

The goal of this work is to identify a model, using patients' phenotypic time series data alone, that would both (i) discover underlying subtypes of individuals based on their disease trajectories as well as (ii) predict future phenotypic values on an individual basis. The ultimate goal is to use this model to inform targeted, and personalized interventions aimed at both reducing the severity of onset of these disorders and the negative outcomes that accompany them, such as suicide and substance abuse. The difficulty in solving this task is that the longitudinal data usually available in existing cohorts is very short and irregularly measured.

In the field of group-based disease trajectory modeling, there are two primary modeling directions stemming from the different fields of model development: those from the field of machine learning primarily employing Gaussian (or some other) stochastic processes and those from the field of statistics/epidemiology mainly in the form of linear/non-linear mixed modeling with structured covariance between time-points [2]. We use each of these two paradigms in application to identify subtypes and predict future internalizing behavior (e.g. fearfulness and social withdrawal), a phenotype which is predictive of anxiety and depressive disorders in adolescence and adulthood [3]. We used longitudinal data from the Maternal Adversity, Vulnerability and Neurodevelopment (MAVAN) project [4] to assess model performance.

## 1.1 Related Literature

Gaussian processes (GPs) have become popular for modeling time-dependent phenomena owing to their applicability to modeling a wide variety of functions and their ability to handle overfitting more directly [5].The intuitive Gaussian output of a GP (providing both an average line of fit and a confidence bound around this line) makes results interpretable and intuitive across fields and computational expertise [6]. GPs have been successfully developed for group-based trajectory modeling such as the Dirichlet Process-Gaussian Process (DP-GP) developed by Hensman et al [7] along with predictive models for health-related outcomes [8,9].

Linear and non-linear mixed modeling share aspects with GPs, such as structured covariance between time points, and can be identical in the case of Gaussian time covariance. This is also a very broad and flexible class of model, considered state-of-the-art for modeling trajectories in epidemiology, biostatistics and statistics [2]. A downside of this approach is that it requires that the number of groups be known a priori. If the number of groups is indeed known, the a priori setting allows the researcher to incorporate their inductive bias into the model. This feature is demonstrated in one such model by Proust-Lima et al [10]. Their latent class mixture model integrates mixture models with mixed effects and longitudinal models across a wide array of longitudinal data types such as Gaussian, count and time-to-event data and integrating structured correlation between time points into the model [10,11]. The subtypes in neuropsychiatric conditions and socio-emotional development are currently not known therefore models such as DP-GP, where the number of clusters are discovered automatically [7] maybe preferable in this particular regard.

## 2 Models

Given the widespread popularity and demonstrated potential accuracy of Gaussian processes when applied to health data [8], we chose the Dirichlet Process-Gaussian Process (DP-GP/Mixture of Gaussian Processes) set forth by Hensman et al [7] as our primary model for simultaneously identifying a generative process and clustering individuals. This model was implemented using a Chinese Restaurant Process (CRP) [12] representation of the Dirichlet Process and Hensman et al's model of the DP-GP [7].

The model is implemented in the following way. For each cluster $k$, we have $Y_k = \{y_n : n \in individuals\ in\ the\ k^{th}\ cluster\}$ and $T_k = \{t_n : n \in individuals\ in\ the\ k^{th}\ cluster\}$. Under this model, each cluster has a latent GP function that governs the time series in that cluster. We denote this function as $f_k(t) \sim GP(0, k_{f_k}(t, t'))$. Given that the individual belongs to this cluster, each data trajectory is modeled as $y_n(t) \sim GP(f_k, k_n(t, t'))$. After marginalizing the latent function, we write a compound covariance function for each cluster as

$$\tilde{k}(t, t', n, n') = \begin{cases} k_n(t, t') + k_{f_k}(t, t') & if\ n = n' \\ k_{f_k}(t, t') & otherwise \end{cases}$$

Thus for each individual in the cluster $k$, we have

$$y(t, n) = GP(0, \tilde{k}(t, t', n, n'))$$

To get the likelihood, we construct $\hat{y} = [y_1^T ... y_{N_k}^T]^T$ and $\hat{t} = [t_1^T ... t_{N_k}^T]^T$. We also construct a covariance matrix $\hat{K}$ using the covariance function and the $\hat{t}$. Then the likelihood for the data is given by

$$p(Y_k|T_k, \theta) = N(\hat{y}|0, \hat{K}).$$

Inference was done using Gibbs sampling and the hyperparameters were chosen using a grid search driven by predictive accuracy relative to the root mean squared error of the prediction.

The LCMM model was taken from [13] as implemented in the R package (lcmm). For completeness, the model is described by Proust-Lima et al as:

$$Y_{ij}|_{c_i=g} = X_{L1i}(t_{ij})^T \beta + X_{L2i}(t_{ij})^T v_g + Z(t_{ij})^T u_i + w_i(t_{ij}) + \epsilon_{ij}$$

where $X_{L1i}(t_{ij})$ and $X_{L2i}(t_{ij})$ describe predictor variables split between shared and class-specific fixed effects $\beta$ and $v_g$, $Z(t_{ij})$ are random effects (set to 0 for our purposes), and $w_i(t_{ij})$ specifies a correlative process between time points. Because this model is implemented in CRAN we leave it to the reader to reference Proust-Lima's paper for a full explanation of the model's derivation [13].

## 3 Experiments

### 3.1 Data

The data used for this project came from the MAVAN cohort, a community based birth cohort from Hamilton, ON and Montreal, QC [4]. Four time points measured on 95 children (48% male) over 3.5 years were used as our longitudinal data in our social-emotional phenotype: internalizing behaviors. Internalizing behaviors relate to disorders such as depression and anxiety, the persistence of which in early childhood has been shown to increase risk of depression later in life [3]. Internalizing behaviors were measured using the Infant Toddler Social Emotional Assessment (ITSEA) questionnaire [17] at ages 1.5 and 2 years and the Child Behavior Checklist (CBCL) [18] at ages 4 and 5 years. These measures were transformed from percentiles to standard normal z-scores across individuals at each time point to better conform to the assumptions of the methods they were modeled with. This data comes with several clear challenges: (1) a relatively small number of longitudinal observations and (2) varied time-spacing between observations (6, 24, and 12 months respectively). We believe these features to be important to our data since often data comes in less than ideal observational frequency and size. Therefore, the behavior of state-of-the-art models on such data is important to explore.

### 3.2 Results

LCMM models were first compared within models with the same time covariance function (no time covariance, autoregressive time (AR) covariance, and Brownian motion (BM) time covariance) to choose the "best" number of groups using BIC. The models from each covariance function with the "best" number of groups was then assessed for predictive accuracy using root mean squared error (RMSE) and correlation between predicted and true values in a held out test set which was made up of 30% of the sample having their final time point removed and then predicted by the remaining data. That is, training was done on the full sample except the final time point observations of 29 individuals and testing was done by predicting these final time points in these individuals and comparing this value to the observed value. Figure 1 shows that the LCMM model with no-time covariance function performed the best of the three LCMM models with no- (NC), autoregressive (AR) and Brownian motion (BM) time covariances. This is evident from Figures 1b and 1c as NC3 (with 3 clusters and the lowest BIC) and NC4 (with 4 clusters) models provide the lowest RMSE and the highest correlation. Interestingly NC4 is a significantly better predictive model than NC3 even though the complexity tradeoff is not in its favor. Comparing NC3, NC4 and DPGP, we find that NC4 outperforms DPGP though 50 trials were not enough to call the difference significant. The difference in running time between the algorithms is substantial - 5min for LCMM for our sample vs 40min for DPGP. Thus, if we were to pick one of these two state-of-the-art frameworks to analyze and build predictive models for the short time series of internalizing behavior, based on the performance we would recommend using LCMM.

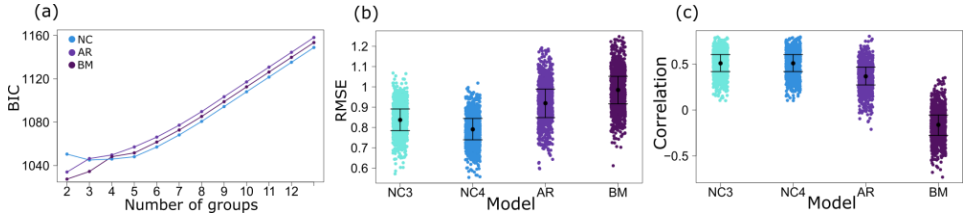

Figure 1: Model selection for LCMM. (a) BIC for three different types of covariance function NC (no-time covariance), AR (autoregressive time covariance), BM (Brownian motion time covariance; NC3 is a 3 cluster model corresponding to the lowest BIC and NC4 corresponding to the next best model (b) RMSE over 200 random held out final points; (c) Pearson Correlation over 200 random held out final points (with mean and inter-quartile range shown in black).

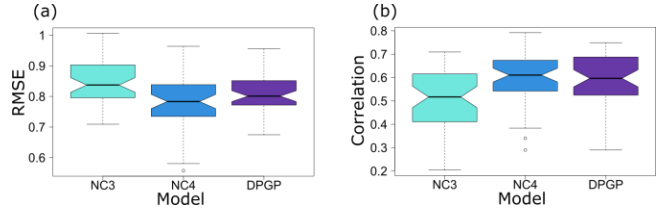

Figure 2: Model comparison: (a) RMSE (b) Correlation. The lower and upper quartiles and the median are based on 50 trial runs.

## 4    Discussion

As researchers coming from the machine learning community, we started with somewhat of a bias towards Gaussian Process models, we expected these models to outperform LCMM, the staple model in epidemiology. However, our particular application and the perils of real data proved us wrong: LCMM turned out to be a faster and a potentially more accurate solution. The large and interacting number of hyperparameters in the DP-GP make optimization difficult with only a few time points. Implementation of DP-GP model with a CRP adds to this constraint by increasing the processing time since the inference is done via Gibbs sampling. Meanwhile, the speed of the LCMM is undone if the number of clusters is not known a priori. The need for testing the model's performance across different numbers of clusters and different covariance structures, which, if it incorporates testing prediction, can substantially increase its processing time.

The results of these analyses point to shortcomings that current state-of-the-art models when applied to longitudinal data under less than ideal circumstances may exhibit. While LCMM might perform slightly better than DPGP, the correlation and RMSE are still high and if we are interested in the translational aspect of these methodologies, much remains to be done. The majority of available socio-emotional longitudinal data during childhood development have relatively small sample size (compared to, for example, electronic medical records data) and each time series is non-uniformly measured over a small number of time points. It is thus imperative that as machine learners we consider these far from ideal scenarios and improve our models to help psychologists and psychiatrists make headway in understanding refining trajectories and improving mental health care.